\documentclass[conference]{IEEEtran}
\IEEEoverridecommandlockouts

\usepackage{cite}
\usepackage{amsmath,amssymb,amsfonts}
\usepackage{algorithmic}
\usepackage{graphicx}
\usepackage{textcomp}
\usepackage{xcolor}
\usepackage{url}
\usepackage{placeins}

\def\BibTeX{{\rm B\kern-.05em{\sc i\kern-.025em b}\kern-.08em
    T\kern-.1667em\lower.7ex\hbox{E}\kern-.125emX}}
\usepackage[compatibility=false]{caption}
\usepackage{subcaption}
\usepackage{enumitem}
\graphicspath{{.}}

\renewcommand{\it}[1]{\textit{#1}}

\newcommand{\bt}[1]{\textbf{#1}}

\newcommand{\hide}[1]{\ignorespaces}

\begin{document}

\title{Similarity-Based Bike Station Expansion via Hybrid Denoising Autoencoders
    \\ \thanks{This research was funded by the PERSEUS project under the Marie Skłodowska-Curie grant agreement No. 101034240.}
}
\author{
    \IEEEauthorblockN{Oluwaleke Yusuf}
    \IEEEauthorblockA{\textit{Department of Engineering Cybernetics} \\
        \textit{Norwegian University of Science} \\ \textit{and Technology (NTNU)} \\
        NO-7491 Trondheim, Norway \\
        ORCID: 0000-0002-5904-648X}
    \and
    \IEEEauthorblockN{M. Tsaqif Wismadi}
    \IEEEauthorblockA{\textit{Department of Architecture and Planning} \\
        \textit{Norwegian University of Science} \\ \textit{and Technology (NTNU)} \\
        NO-7491 Trondheim, Norway \\
        ORCID: 0009-0001-7385-896X}
    \and
    \IEEEauthorblockN{Adil Rasheed}
    \IEEEauthorblockA{\textit{Department of Engineering Cybernetics} \\
        \textit{Norwegian University of Science} \\ \textit{and Technology (NTNU)} \\
        NO-7491 Trondheim, Norway \\
        ORCID: 0000-0003-2690-983X}
}
\maketitle

\begin{abstract}
    Urban bike-sharing systems require strategic station expansion to meet growing demand. Traditional allocation approaches rely on explicit demand modelling that may not capture the urban characteristics distinguishing successful stations. This study addresses the need to exploit patterns from existing stations to inform expansion decisions, particularly in data-constrained environments.
    We present a data-driven framework leveraging existing stations deemed desirable by operational metrics. A hybrid denoising autoencoder (HDAE) learns compressed latent representations from multi-source grid-level features (socio-demographic, built environment, and transport network), with a supervised classification head regularising the embedding space structure. Expansion candidates are selected via greedy allocation with spatial constraints based on latent-space similarity to existing stations.
    Evaluation on Trondheim's bike-sharing network demonstrates that HDAE embeddings yield more spatially coherent clusters and allocation patterns than raw features. Sensitivity analyses across similarity methods and distance metrics confirm robustness. A consensus-based procedure across multiple parametrisations distils 32 high-confidence extension zones where all parametrisations agree.
    The results demonstrate how representation learning captures complex patterns that raw features miss, enabling evidence-based expansion planning without explicit demand modelling. The consensus procedure strengthens recommendations by requiring agreement across parametrisations, while framework configurability allows planners to incorporate operational knowledge. The methodology generalises to any location-allocation problem where existing desirable instances inform the selection of new candidates.
\end{abstract}

%%%%%=============================================================================%%%%

\section{Introduction}
\label{sec:introduction}
Bike-sharing systems (BSS) have emerged as critical components of sustainable urban mobility, providing flexible options that complement public transit networks for first-mile/last-mile connectivity and short-distance trips. As cities expand and demand for active mobility increases, BSS operators face the challenge of strategically placing new stations to maximise accessibility and utilisation while maintaining operational efficiency. The location-allocation problem in BSS planning is particularly complex, as successful station placement depends on intricate interactions between urban form, transport infrastructure, socio-demographic patterns, and user behaviour.

Traditional approaches to BSS expansion employ explicit demand modelling combined with optimisation frameworks for facility location problems such as Weighted Linear Combination \cite{Malczewski2000otu} or Maximal Coverage Location Problem \cite{Church1974tmc}. However, such approaches often require detailed demand data that may not be available in expanding systems or emerging markets. Furthermore, mathematically rigorous solutions are not guaranteed to fully capture the nuanced urban characteristics that distinguish successful station locations from other alternatives. An alternative paradigm is to leverage patterns from existing successful stations to inform expansion decisions through a similarity-based approach that assumes locations sharing key features with existing stations are likely to perform well.

The core challenge addressed in this study is developing a framework for identifying suitable expansion locations for bike-sharing stations in a purely data-driven manner that exploits existing stations deemed desirable or successful by whatever expert metric. Such a framework avoids imposing value judgements on what constitutes a ``good'' location, instead seeking to find candidate locations exhibiting similar urban characteristics to existing stations under the assumption that such similarity predicts operational success.

However, urban environments are characterised by high-dimensional, heterogeneous data spanning socio-demographics, built environment, terrain, mobility patterns and transit accessibility. Directly computing similarity in this raw feature space presents challenges related to noise, redundancy, and the curse of dimensionality. In addition, the number of existing stations is vastly smaller than the pool of candidate locations, creating severe class imbalance. Finally, spatial constraints (such as minimum separation distances) must be enforced to avoid cannibalisation of nearby stations. An effective solution must learn compact representations that capture the essential characteristics distinguishing existing and potential station locations while remaining robust to noise and capable of producing interpretable, spatially coherent expansion plans.

This paper proposes a framework based on hybrid denoising autoencoders (HDAE) to address these challenges. The primary objective is to develop a representation learning approach that compresses high-dimensional urban features into a structured latent space where similarity-based candidate selection becomes tractable and effective. The framework comprises five key components: \it{(i)} spatial tessellation and multi-source feature engineering to create comprehensive urban descriptors, \it{(ii)} hybrid denoising autoencoder training that combines reconstruction objectives with supervised classification to regularise latent representations, \it{(iii)} similarity computation using multiple methods and distance metrics within the learned embedding space, \it{(iv)} greedy allocation algorithms incorporating spatial proximity constraints to produce candidate station sets, and \it{(v)} consensus-based extension selection that pools candidates across multiple parametrisations and retains only unanimously agreed locations.

The methodology addresses fundamental questions about the trade-offs between raw features and learned embeddings, the effectiveness of different similarity measures, and the sensitivity of selection outcomes to key parameters. By comparing HDAE-based allocation outcomes against those from raw features, we evaluate whether the added modelling complexity yields meaningful improvements in candidate coherence and spatial structure. Furthermore, the framework is designed to be generalisable beyond BSS expansion to other location-allocation problems where existing desirable or successful instances can be used to inform selection of new candidates based on the similarity of their underlying characteristics.

The primary contributions of this study can be summarised as follows:

\begin{itemize}
    \item A data-driven framework for BSS expansion that leverages hybrid denoising autoencoders to learn structured latent representations of urban characteristics, enabling similarity-based candidate selection without explicit demand modelling.
    \item Comparative analysis demonstrating that HDAE embeddings produce more coherent spatial patterns and improved clustering quality compared to raw features, validating the representation learning approach.
    \item Comprehensive evaluation of similarity measures (top-$k$, kernel density estimation) and distance metrics (cosine, Euclidean) for candidate selection, including sensitivity analysis of key parameters.
    \item A consensus-based extension selection procedure that pools candidates across multiple parametrisations and retains only unanimously agreed locations, resolving parameter uncertainty without ad-hoc tuning.
\end{itemize}

%%%%%=============================================================================%%%%

\section{Methodology}
\label{sec:methodology}

This section describes the HDAE-based framework for similarity-based BSS expansion. The approach integrates spatial feature engineering, representation learning via hybrid denoising autoencoders, and greedy allocation with spatial constraints. Finally, a consensus mechanism across multiple similarity parametrisations yields robust station extension recommendations.

\subsection{Study Area and Feature Engineering}

The methodology is demonstrated on Trondheim, Norway, a mid-sized city with an operational bike-sharing system (Trondheim Bysykkel). The study area encompasses the urban core and surrounding residential neighbourhoods, discretised into a regular grid of $100 \times 100$ metre cells. This yields 19,474 grid cells, of which 68 contain existing BSS stations (noting that individual cells may contain multiple station access points).

\begin{figure*}[tb!]
    \centering \includegraphics[width=\linewidth]{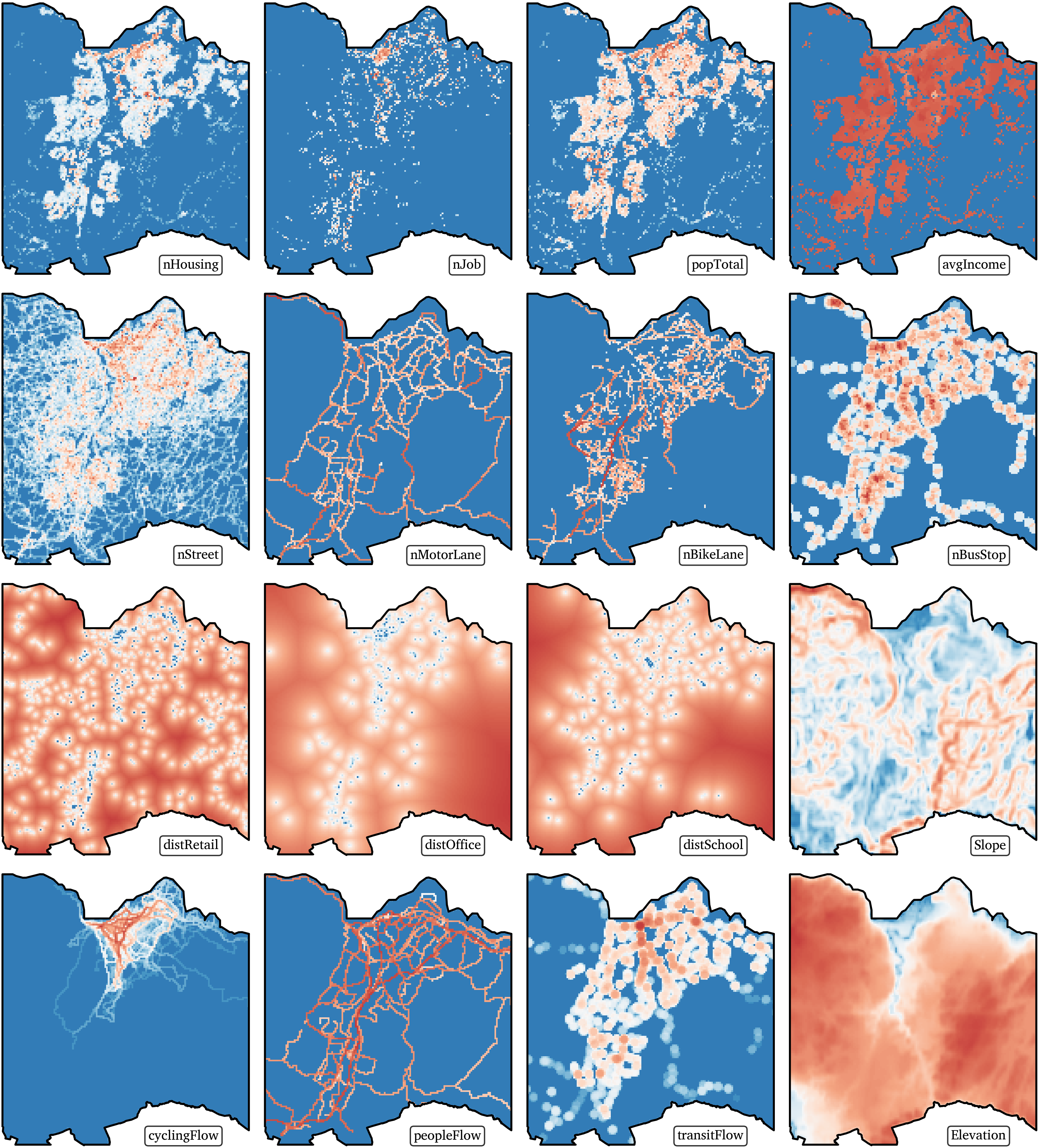}
    \caption{Spatial distribution of selected engineered features (16 of 29) across the study area. \it{Note: All values log-transformed for visualisation.}}
    \label{fig:feature-heatmap}
\end{figure*}

Spatial datasets from multiple sources are integrated to create comprehensive urban descriptors for each grid cell. The raw data is aggregated and transformed via standard GIS operations: point-in-polygon for counting features within cells, intersection for linear features (e.g., road networks), and centroid-based distance calculations for proximity measures. Transit features employ a 250m catchment radius, with distance-weighted allocation of boarding and alighting counts from nearby stops.
Fig.~\ref{fig:feature-heatmap} presents the spatial distribution of representative features across the study area, with values log-transformed for visual clarity. The final feature set comprises 29 engineered variables organised into four thematic groups:

\begin{itemize}
    \item \it{Socio-Demographic Features} capture residential intensity, employment density, and socioeconomic context through population, housing units, jobs, and average income derived from census-based grid data.
    \item \it{Built Environment Features} quantify urban form and land use via building footprints (retail, offices, schools), proximity to the central business district, and terrain characteristics (slope, elevation) that constrain cycling accessibility.
    \item \it{Transport Network Features} encompass cycling infrastructure (cycleway and bike lane presence), road network connectivity (street segments, junctions, motor lanes), mobility flows from GPS and cellular data, and public transit accessibility (stop counts, line diversity, stop proximity, and passenger activity).
    \item \it{Neighbourhood Flow Features} smooth the spatial landscape through Moore neighbourhood aggregation (Chebyshev distance $\leq 2$ hops), computing mean and maximum values for cycling, people, and transit flows across neighbouring grid cells to capture local spatial context.
\end{itemize}

\subsection{Hybrid Denoising Autoencoder}

\begin{figure}[tb!]
    \centering \includegraphics[width=0.90\linewidth]{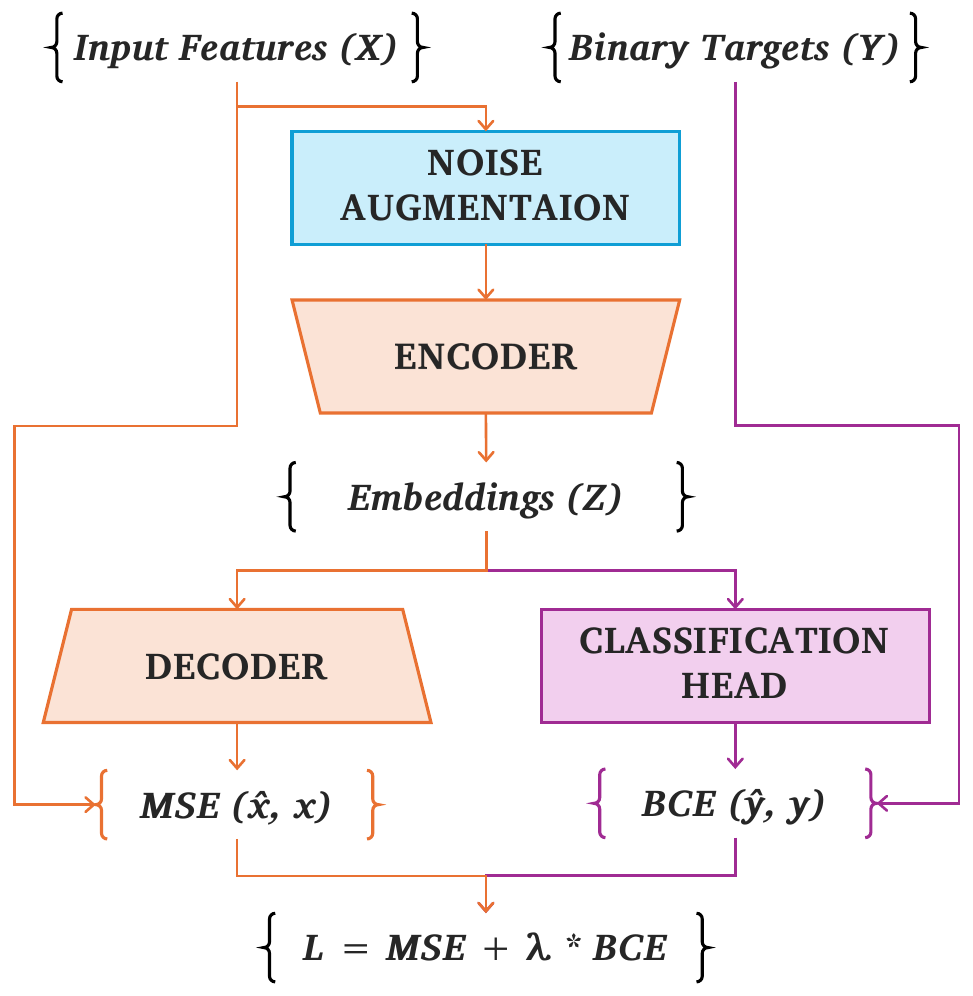}
    \caption{Hybrid denoising autoencoder (HDAE) architecture with four components: encoder (compression), noise augmentation (robustness), decoder (reconstruction), and classification head (latent regularisation).}
    \label{fig:hdae-architecture}
\end{figure}

The similarity-based pipeline can work with raw z-scored features, but computing similarity in a 29-dimensional space presents challenges: noise, redundancy, and the curse of dimensionality may obscure meaningful patterns. We employ a hybrid denoising autoencoder (HDAE) \cite{Vincent2010sda} to address these concerns, extending the base autoencoder architecture with two key components: feature noise augmentation and a supervised classification head. Fig.~\ref{fig:hdae-architecture} depicts the complete architecture.

\it{Why Autoencoder?}
This unsupervised representation learning model compresses input features into a lower-dimensional latent representation (\it{encoder}) and reconstructs original features from this representation (\it{decoder}). By training to reconstruct inputs, the encoder learns to capture the most important patterns while discarding noise and irrelevant information. This produces a more compact and informative representation that improves similarity computation quality. The lower-dimensional embeddings also reduce computational cost.

\it{Why Denoising?}
During training, Gaussian noise augmentation is injected into a random subset of features, and the model learns to reconstruct the original clean inputs. This forces the encoder to learn robust representations that remain stable under input perturbations, leading to improved generalisation and clustering quality in the latent space. Thus, grids with similar urban characteristics will cluster together in the embedding space and yield high similarity scores despite minor variations in their raw features.

\it{Why Hybrid?}
A supervised classification head atop the encoded representation forces structure on the embedding space. The classification head must distinguish existing station grids from candidates based on their embeddings. This forces the model to partition the embedding space, but the reconstruction loss resists this separation for those candidates with similar features. The resulting tension places promising candidate grids near the decision boundary and near existing station embeddings, optimistically yielding high downstream similarity scores.

\vspace{0.5\baselineskip}
\subsubsection{Encoder-Decoder Structure}

The base autoencoder architecture employs a symmetric encoder-decoder design with configurable depth. The encoder maps input $\mathbf{x} \in \mathbb{R}^{29}$ to a latent embedding $\mathbf{z} \in \mathbb{R}^{d}$, and the decoder reconstructs $\hat{\mathbf{x}}$ from $\mathbf{z}$. Intermediate layer widths are determined by exponential scaling ($2^i \cdot h$, for base dimension $h$ and layer index $i$). For the two-layer configuration, the encoder comprises three linear transformations:

\begin{equation}
    \begin{aligned}
        \mathbf{h}_1 & = \text{ReLU}(\mathbf{W}_1 \mathbf{x} + \mathbf{b}_1)                                \\
        \mathbf{h}_2 & = \text{ReLU}(\mathbf{W}_2 \mathbf{h}_1 + \mathbf{b}_2)                              \\
        \mathbf{z}   & = f_{\text{enc}}(\mathbf{x}) = \text{ReLU}(\mathbf{W}_3 \mathbf{h}_2 + \mathbf{b}_3)
    \end{aligned}
\end{equation}

\noindent where $\mathbf{h}_1, \mathbf{h}_2$ are intermediate hidden representations. Kaiming uniform initialisation \cite{He2015ddi} and layer normalisation \cite{Ba2016ln} are applied to stabilise training dynamics, with ReLU activations between layers.

\vspace{0.5\baselineskip}
\subsubsection{Denoising Augmentation}

During training, a feature noise layer injects Gaussian noise into a random subset of features, transforming the input $\mathbf{x}$ into a noisy version $\tilde{\mathbf{x}}$:

\vspace{-0.5\baselineskip}
\begin{equation}
    \tilde{\mathbf{x}} = \mathbf{x} + \mathbf{\epsilon} \odot \mathbf{m}, \quad \mathbf{\epsilon} \sim \mathcal{N}(0, \sigma^2), \quad \mathbf{m} \sim \text{Bernoulli}(p)
\end{equation}

\noindent where $\mathbf{m}$ is a binary mask with probability $p=0.3$ of selecting each feature, and $\sigma=0.1$ controls noise magnitude. Noise values are clipped to $[-3, 3]$ standard deviations to prevent extreme perturbations.

\vspace{0.5\baselineskip}
\subsubsection{Classification Head}

The hybrid variant extends the base architecture with a supervised classification head that maps the latent embedding $\mathbf{z}$ from the encoder to a binary logit $\hat{y}$ indicating station presence:

\begin{equation}
    \hat{y} = \mathbf{w}_2^T \cdot \text{ReLU}(\mathbf{W}_1 \cdot \ell_2(\mathbf{z}) + \mathbf{b}_1) + \mathbf{b}_2
\end{equation}

\noindent where $\ell_2(\mathbf{z}) = \mathbf{z} / \|\mathbf{z}\|_2$ normalises embeddings to unit length. The head maps from dimension $d$ through a hidden layer of size $h$ to a single output, with ReLU activation between the layers. The raw logit $\hat{y}$ is passed through a sigmoid activation to produce a probability estimate.

The internal L2 normalisation in the classification head encodes class-discriminative signal in the angular geometry of the latent space rather than in vector magnitudes. This design choice aligns with the intended downstream similarity metric (cosine similarity), which measures directional alignment in the latent space, is insensitive to embedding magnitude, and is well-conditioned in high-dimensional spaces.

The classification head creates dual pressure on the latent space: the classification loss drives separation between station grids and candidates, while the reconstruction loss resists this for feature-similar candidates. The resulting tension places such candidates near the decision boundary and, consequently, near existing station embeddings. \hide{Optimistically, this places promising candidate grids near existing station embeddings, yielding high downstream similarity scores.} Note that the classification head serves solely as a latent space regulariser during training. Its logits are discarded after training, and only the encoder embeddings are used for downstream similarity computation.

\vspace{0.5\baselineskip}
\subsubsection{Loss Function}

The hybrid loss combines reconstruction and classification objectives using a weighted sum of the mean squared error (MSE) for reconstruction and binary cross-entropy (BCE) for classification:

\begin{equation}
    \begin{aligned}
        \mathcal{L}_{\text{MSE}} & = \frac{1}{N} \sum_{i=1}^{N} \|\mathbf{x}_i - \hat{\mathbf{x}}_i\|^2                                \\
        \mathcal{L}_{\text{BCE}} & = -\frac{1}{N} \sum_{i=1}^{N} \left[ \alpha y_i \log(\hat{y}_i) + (1-y_i) \log(1-\hat{y}_i) \right] \\
        \mathcal{L}              & = \mathcal{L}_{\text{MSE}} + \lambda \cdot \mathcal{L}_{\text{BCE}}
    \end{aligned}
\end{equation}

\noindent where $\alpha$ is a class weight applied to the positive class (station presence) to counteract the extreme class imbalance between existing station grids and candidate grids. The parameter $\lambda$ controls the relative importance of the classification loss, ensuring that reconstruction remains the dominant objective while still providing sufficient regularisation to shape the latent space geometry.

\subsection{Data Preprocessing and Model Training}
Due to the disparate scales and units of the engineered features, the combined dataset undergoes z-score normalisation to ensure that each feature contributes proportionately to similarity calculations. The binary target variable indicating station presence is created for supervised classification during HDAE training but is not used in similarity computations.
The dataset is split into training (80\%) and validation (20\%) subsets, with input features organised into a matrix $\mathbf{X} \in \mathbb{R}^{N \times 29}$, where $N$ is the number of grid cells, and the target variable forming a binary vector $\mathbf{y} \in \{0,1\}^N$ indicating station presence.

Due to the extreme class imbalance between existing station grids and candidate grids ($\approx 1:285$), the positive class is up-weighted during loss computation. This approach avoids the need for weighted sampling techniques that could distort the natural spatial distribution for reconstruction objectives.
Furthermore, the HDAE classification head operates on L2-normalised embeddings during training, shaping the latent space geometry under this normalisation. This dual normalisation strategy, z-scoring of inputs and L2-normalisation of embeddings, ensures well-conditioned similarity calculations while maintaining reconstruction quality.

Hyperparameter optimisation sweeps explore configurations varying latent dimension $d \in \{8, 12, 16\}$, classification weight $\lambda \in [0.1, 0.5]$, and class weight $\alpha \in [1, 50]$. The selected configuration uses a two-layer encoder-decoder ($29 \to 32 \to 16 \to 8$) with $d=8$, $\lambda=0.1$, and $\alpha=10$.
The model was trained using the Adam optimiser \cite{Kingma2017aam} at a learning rate of $10^{-3}$ and batch size of 512. Training ran for up to 1024 epochs, with early stopping based on validation loss improvement (patience of 15 epochs) to prevent overfitting.

\subsection{Similarity-Based Location-Allocation Pipeline}

Following HDAE training, each grid cell's latent embedding is paired with its spatial location, enabling similarity-based allocation constrained by proximity. The pipeline comprises three stages: conflict graph construction, similarity computation, and greedy selection with local search refinement.

\vspace{0.5\baselineskip}
\subsubsection{Conflict Graph Construction}

A spatial conflict graph encodes proximity constraints among candidate locations. The algorithm first identifies existing station grids and candidate grids (all others). Candidates are filtered using a BallTree spatial index to exclude those within a buffer radius of 250m from any existing station. This ensures that new stations are not placed too close to existing ones, preventing cannibalisation and promoting spatial coverage.

Among the remaining candidates, edges connect grid pairs whose centroids lie within the buffer distance, forming a conflict graph stored as an adjacency list. Independent sets in this graph represent spatially dispersed selections where no two selected grids are closer than the buffer threshold. This graph structure is used during the allocation process to enforce similar spatial constraints among selected candidates.

\vspace{0.5\baselineskip}
\subsubsection{Similarity Computation}

The similarity weights quantify how well each candidate grid matches existing station grids in the latent space. In this study, we use all existing station grids as the reference set $R$ (which existing station grids to compare against) and all non-station grids within the study area as the candidate set $C$ (which candidate grid cells to consider). However, the framework is designed to be flexible such that the reference and candidate sets can be configured based on operational knowledge or specific expansion scenarios. For example, an operator could restrict $R$ to specific high-performing stations or limit $C$ to a target expansion zone.

The similarity computation depends on two independent choices: the aggregation method (which reference grids to combine) and the distance metric (how to measure distances in the embedding space). Let $\mathbf{z}_{c,i}$ and $\mathbf{z}_{r,j}$ denote the embeddings of candidate $c_i$ and reference $r_j$ respectively.

\vspace{0.25\baselineskip}
\begin{itemize}[itemsep=0.5\baselineskip]
    \item \bt{Method}: Two aggregation methods are explored, each combining pairwise similarities into a single weight per candidate:

          \begin{itemize}
              \item \it{Top-$k$}: The weight is the mean of top-$k$ similarities to $R$ in embedding space. $k=1$ reduces to the nearest-neighbour case (maximum similarity to any existing station) while higher $k$ values capture broader alignment with $R$ rather than outliers.
                    \begin{equation}
                        w_i = \frac{1}{k} \sum_{j \in \text{top-}k} d(\mathbf{z}_{c,i}, \mathbf{z}_{r,j})
                    \end{equation}
                    $k=3$ is chosen as the default for a balance between local and global similarity.

              \item \it{KDE}: The weight is a radial basis function kernel density estimate that sums contributions from the entire $R$, with closer grids (in embedding space) contributing more to the weight.
                    \begin{equation}
                        w_i = \sum_{j \in R} \exp\left(-\frac{d(\mathbf{z}_{c,i}, \mathbf{z}_{r,j})^2}{2\sigma^2}\right)
                    \end{equation}
                    $\sigma$ controls the kernel bandwidth, computed as the median of pairwise distances between candidates and reference grids.
          \end{itemize}

    \item \bt{Metric}: Two distance metrics are evaluated for measuring pairwise similarity between the candidate and reference embeddings $d(\mathbf{z}_{c,i}, \mathbf{z}_{r,j})$, negated to convert distance into similarity for maximisation:

          \begin{itemize}
              \item \it{Cosine}: $d(\cdot, \cdot) = -(\mathbf{z}_{c,i} \cdot \mathbf{z}_{r,j}) / (\|\mathbf{z}_{c,i}\| \|\mathbf{z}_{r,j}\|)$. This requires z-scored and L2-normalised embeddings.

              \item \it{Euclidean}: $d(\cdot, \cdot) = -\|\mathbf{z}_{c,i} - \mathbf{z}_{r,j}\|_2^2$. This only requires z-scored embeddings.
          \end{itemize}
\end{itemize}

\begin{figure*}[tb!]
    \centering
    \begin{subfigure}{0.49\linewidth}
        \includegraphics[width=\linewidth]{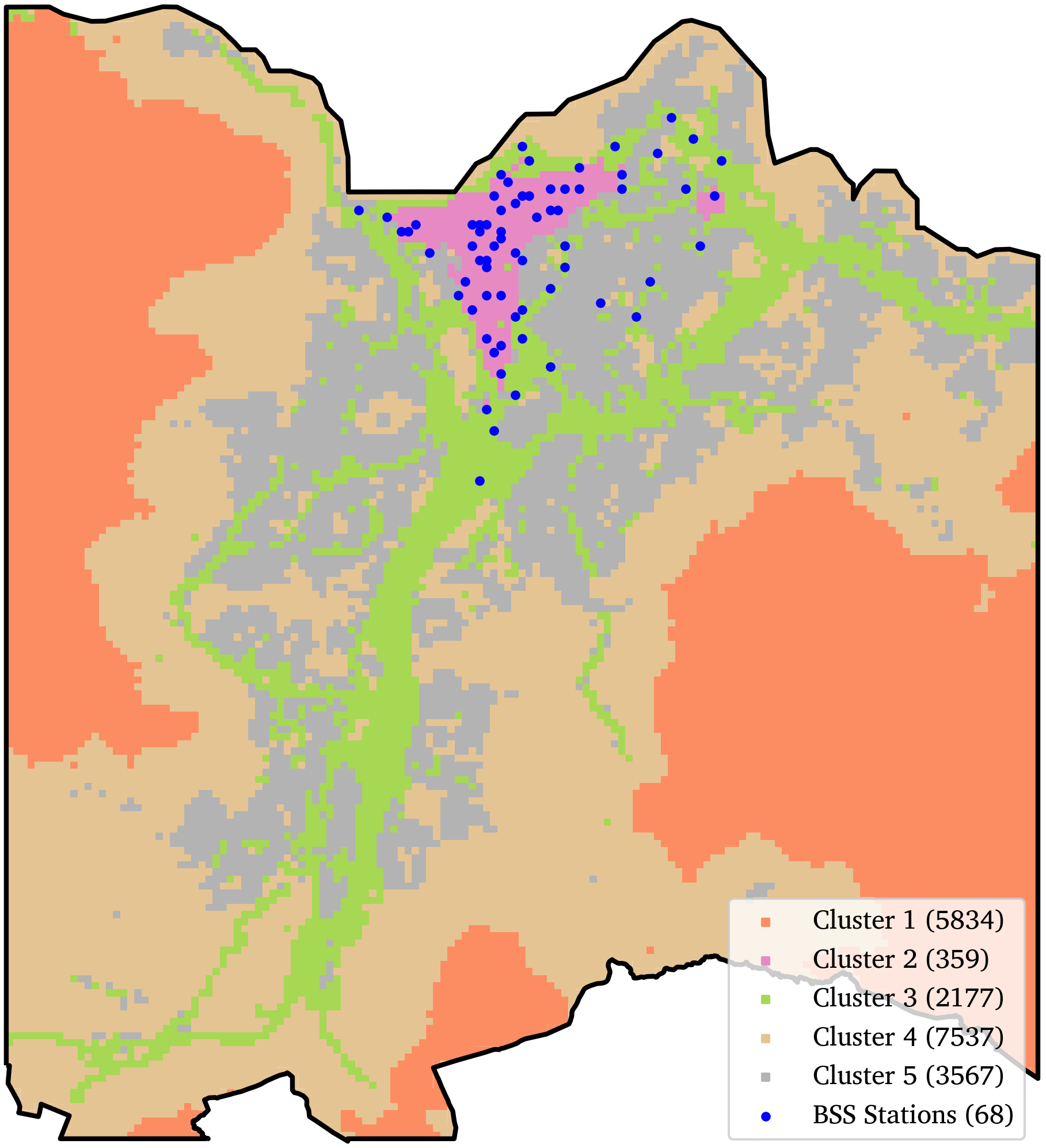}
        \caption{Raw Features}
    \end{subfigure}
    \begin{subfigure}{0.49\linewidth}
        \includegraphics[width=\linewidth]{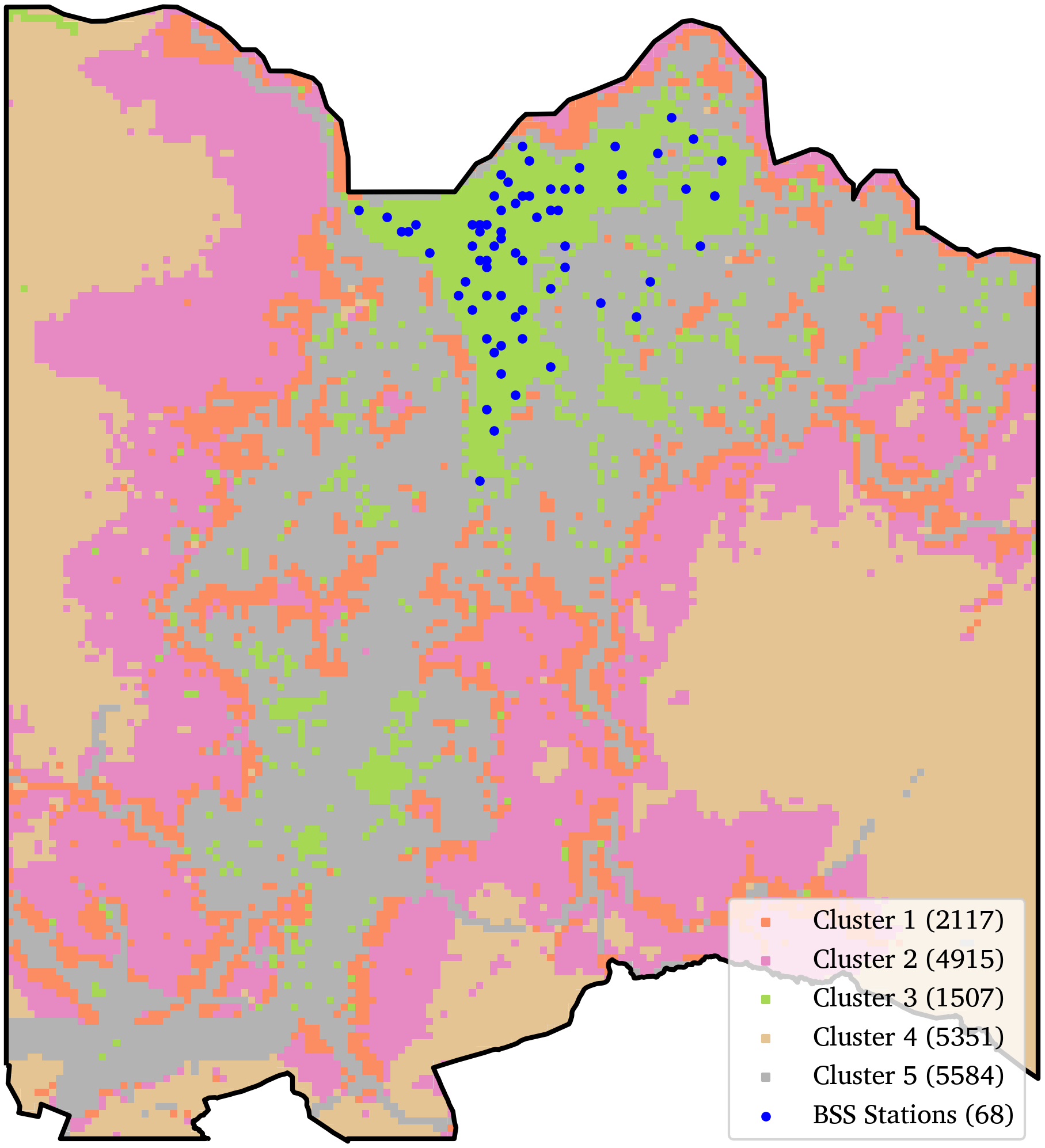}
        \caption{HDAE Embeddings}
    \end{subfigure}
    \caption{Spatial cluster maps ($k=5$) comparing raw features (left) versus HDAE embeddings (right). HDAE embeddings produce finer, heterogeneous clusters with intermixed pockets reflecting richer urban distinctions.}
    \label{fig:clustering-comparison}
\end{figure*}

\vspace{0.5\baselineskip}
\subsubsection{Greedy Selection with Local Search}

The allocation algorithm implements a greedy maximum-weight independent set (MWIS) heuristic on the conflict graph to enforce the spatial constraints. Candidate grids are ranked by similarity weight in descending order. Iteratively, the highest-weight available candidate is selected, and its neighbours in the conflict graph are removed from consideration. This continues until the desired $n$ candidates are selected or no more candidates remain.

Following greedy selection, swap-based local search explores improvements. For each selected candidate, the algorithm considers swapping with non-selected candidates outside its conflict neighbourhood, accepting swaps that increase total weight across all $n$ selections. The solution is refined over multiple iterations to escape local optima potentially induced by greedy ordering. The final output is a set of $n$ candidate grid cells that are spatially dispersed (due to conflict graph constraints) and similar to existing stations in the latent space (due to similarity-based weighting).

%%%%%=============================================================================%%%%

\section{Results}
\label{sec:results}

The HDAE framework is evaluated on Trondheim's bike-sharing network by comparing embedding quality and allocation outcomes against raw features, examining the learned latent space, assessing sensitivity to key parameters, and deriving a consensus set of high-confidence extension zones.

\subsection{Comparison of Embeddings vs. Raw Features}

To evaluate embedding quality, clustering analysis compares HDAE representations against raw features. Fig.~\ref{fig:clustering-comparison} presents spatial cluster maps produced by $k$-means clustering ($k=5$) applied to the raw feature and HDAE embeddings, each z-scored prior to clustering. The raw feature clusters largely form non-overlapping layers radiating inward from the study area boundary towards the city centre, with the exception of one cluster that follows the major road network. This suggests the raw features mainly capture how far a location is from the city centre, rather than the specific mix of urban characteristics at each location.

In contrast, HDAE embeddings yield spatially heterogeneous clusters in which pockets of one cluster appear within another, reflecting finer distinctions in urban character that the raw features cannot resolve. The cluster that covers the city centre is more extensive, covers all existing station grids, and includes Trondheim's ``second centre'' to the south (Heimdal) and the large residential and educational hub to the east (Moholt). This qualitative observation is confirmed quantitatively: the mean silhouette score increases from 0.135 for raw features to 0.253 for HDAE embeddings, indicating substantially tighter within-cluster cohesion and clearer between-cluster separation in the learned latent space.

\begin{figure}[tb!]
    \centering \includegraphics[width=\linewidth]{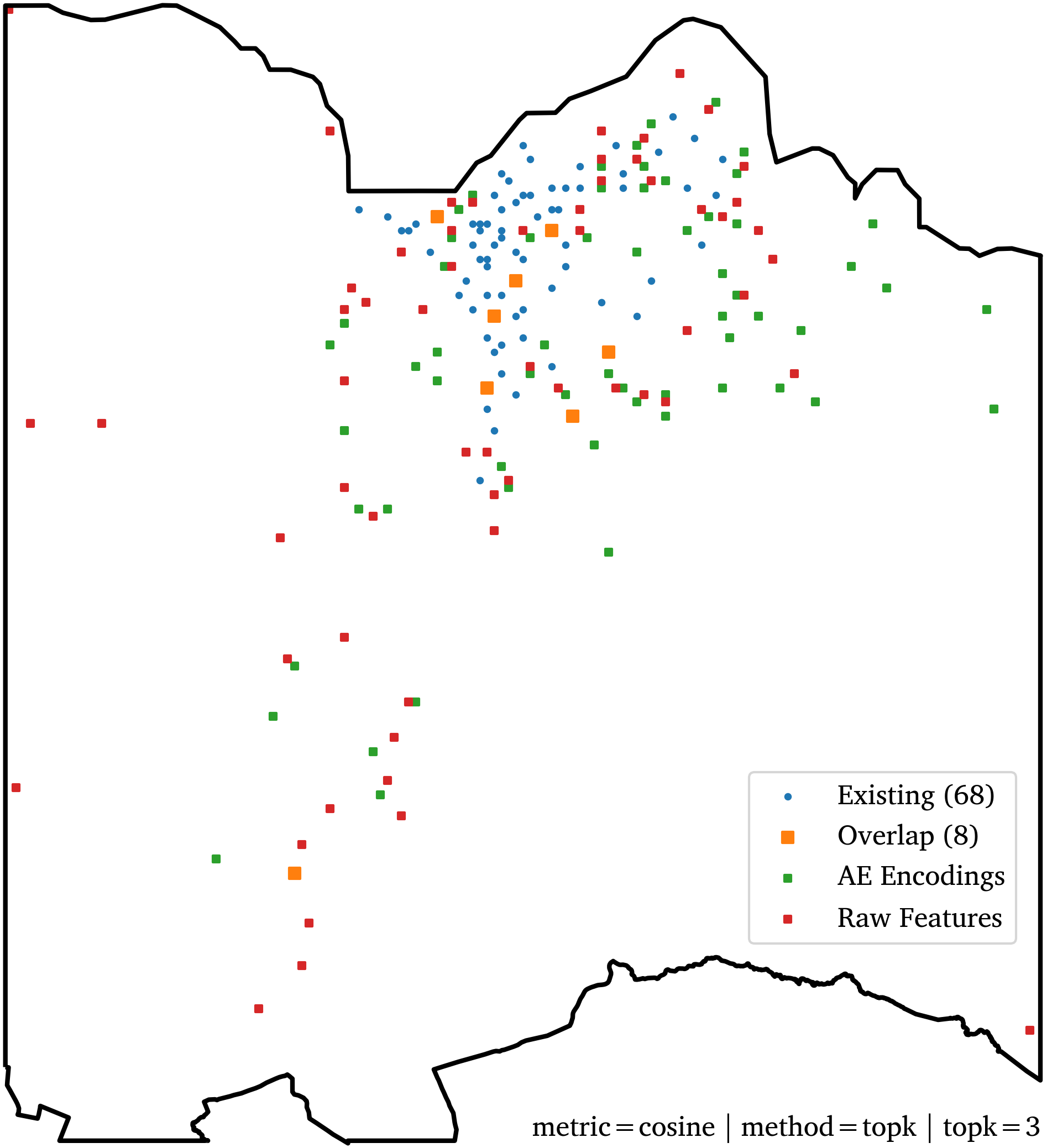}
    \caption{Comparison of allocation outcomes between HDAE embeddings and raw features, selecting 68 candidates with a 250m buffer.}
    \label{fig:allocation-results}
\end{figure}

The same contrast emerges in allocation outcomes between raw features and HDAE embeddings. Fig.~\ref{fig:allocation-results} compares the spatial distribution of $n=68$ selected candidates using reasonable default parameters: top-$k$ method with $k=3$, cosine metric, and 250m proximity buffer. The raw features result in scattered noise-driven selections at the periphery of the study area, influenced by isolated land-use signals that are unlikely to support viable BSS stations.
HDAE-based allocation identifies candidates within well-developed urban areas sharing meaningful characteristics with the reference grids, consistent with its more informative cluster structure. Only 8 of the 68 candidates (11.76\%) are selected by both raw features and HDAE embeddings, indicating that the learned representations capture patterns that the raw features miss, leading to substantially different allocation outcomes.

\subsection{Embedding Analysis}

\begin{figure}[tb!]
    \centering \includegraphics[width=0.95\linewidth]{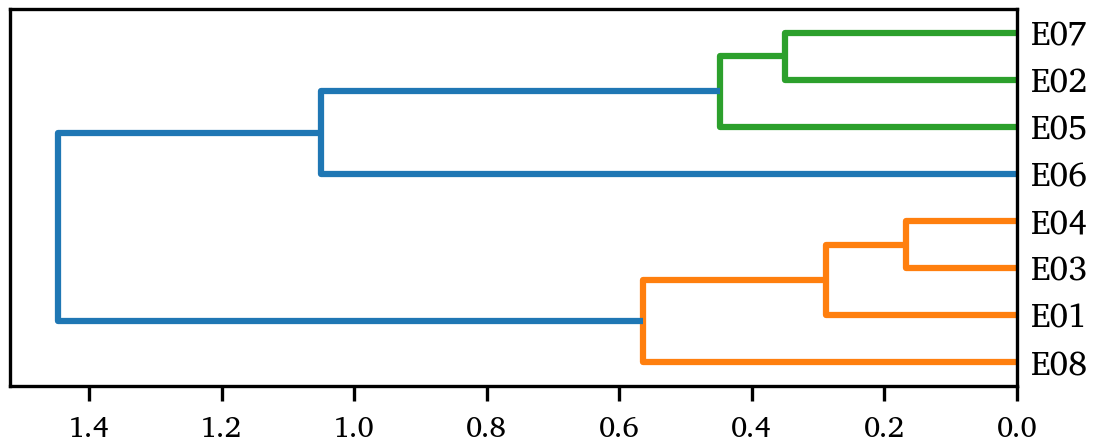}
    \caption{Pairwise correlation between HDAE embedding dimensions, showing low inter-dimension correlations and effective distribution of information across the latent space.}
    \label{fig:embedding-correlation}
\end{figure}

To verify that the latent dimensions capture diverse aspects of the engineered spatial features, Fig.~\ref{fig:embedding-correlation} shows the pairwise correlations between the eight embedding dimensions. The low inter-dimension correlation indicates that the autoencoder effectively captures diverse, non-redundant aspects of the input features and distributes the information across the latent space.

\begin{figure}[tb!]
    \centering \includegraphics[width=\linewidth]{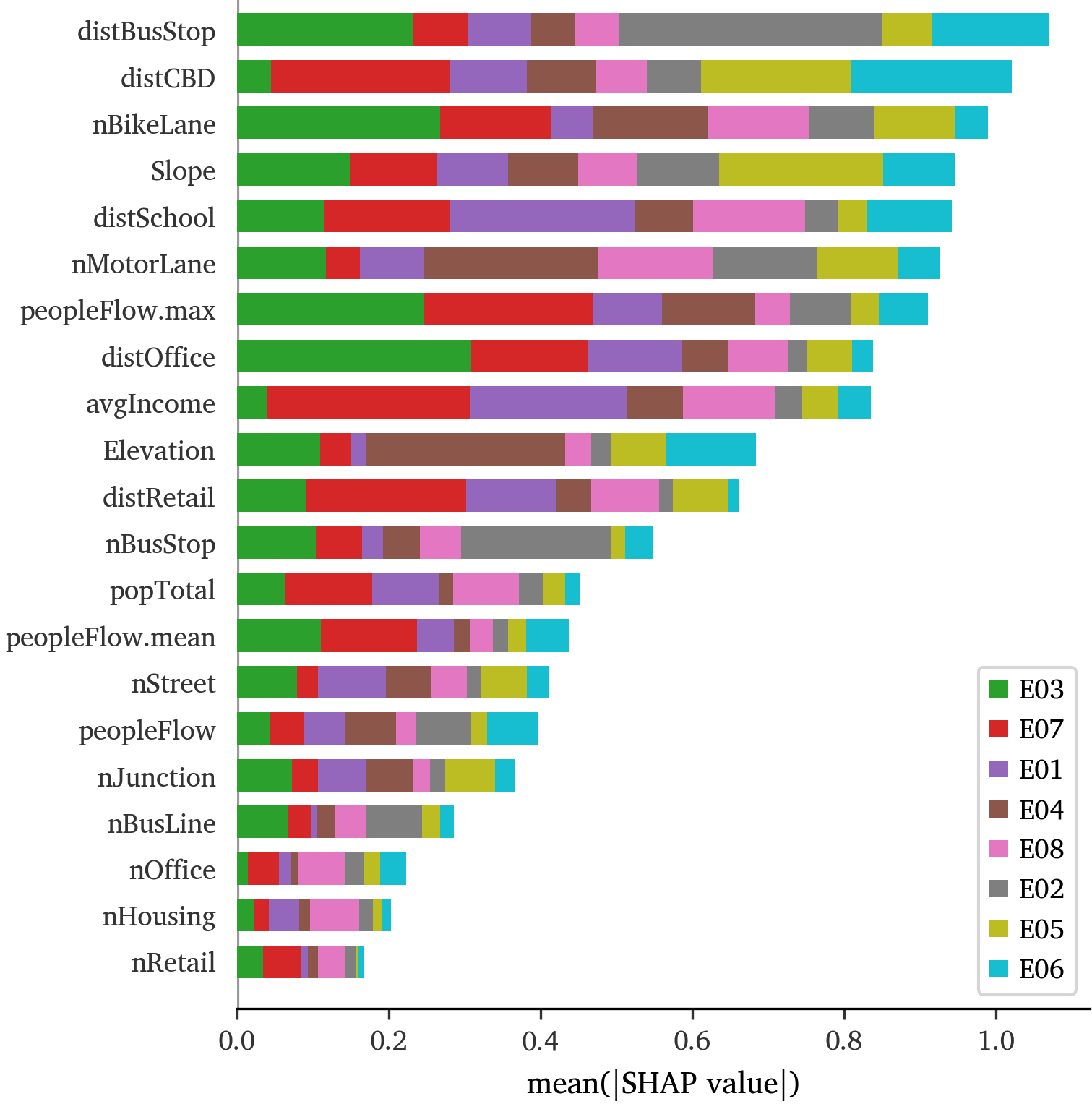}
    \caption{SHAP summary plot showing the top-21 ranked spatial features and their contribution to the HDAE latent dimensions.}
    \label{fig:shap-analysis}
\end{figure}

SHAP (SHapley Additive exPlanations) analysis \cite{Lundberg2017aua} identifies which raw features most influence the latent embeddings and how they distribute across dimensions. Fig.~\ref{fig:shap-analysis} shows feature importance for the top-21 ranked spatial features. The highest-ranked features are split between socio-demographic and transport network groups, with distance to nearest transit stop, distance to city centre, and number of bike lanes ranked at the top. The transit and cycling flow features rank lowest, suggesting the model captured their information in the people flow feature (and its neighbourhood aggregation) which captures overall crowd movement regardless of mobility mode.

In addition, the importance of features is distributed across the embedding dimensions, with no single feature dominating any particular dimension. This suggests that the model has learned to capture complex interactions between features rather than relying on a few dominant signals, contributing to the improved clustering and allocation performance observed with HDAE embeddings.

\subsection{Sensitivity Analysis}

To assess how robust the allocation outcomes are to parameter choices, results are compared across similarity aggregation methods, distance metrics, and their key parameters. For all allocations, the same trained HDAE embeddings are used to select $n=68$ with a 250m proximity buffer. The existing station grids are shown in the relevant figures for context.

\begin{figure*}[tb!]
    \centering
    \begin{subfigure}{0.49\linewidth}
        \includegraphics[width=\linewidth]{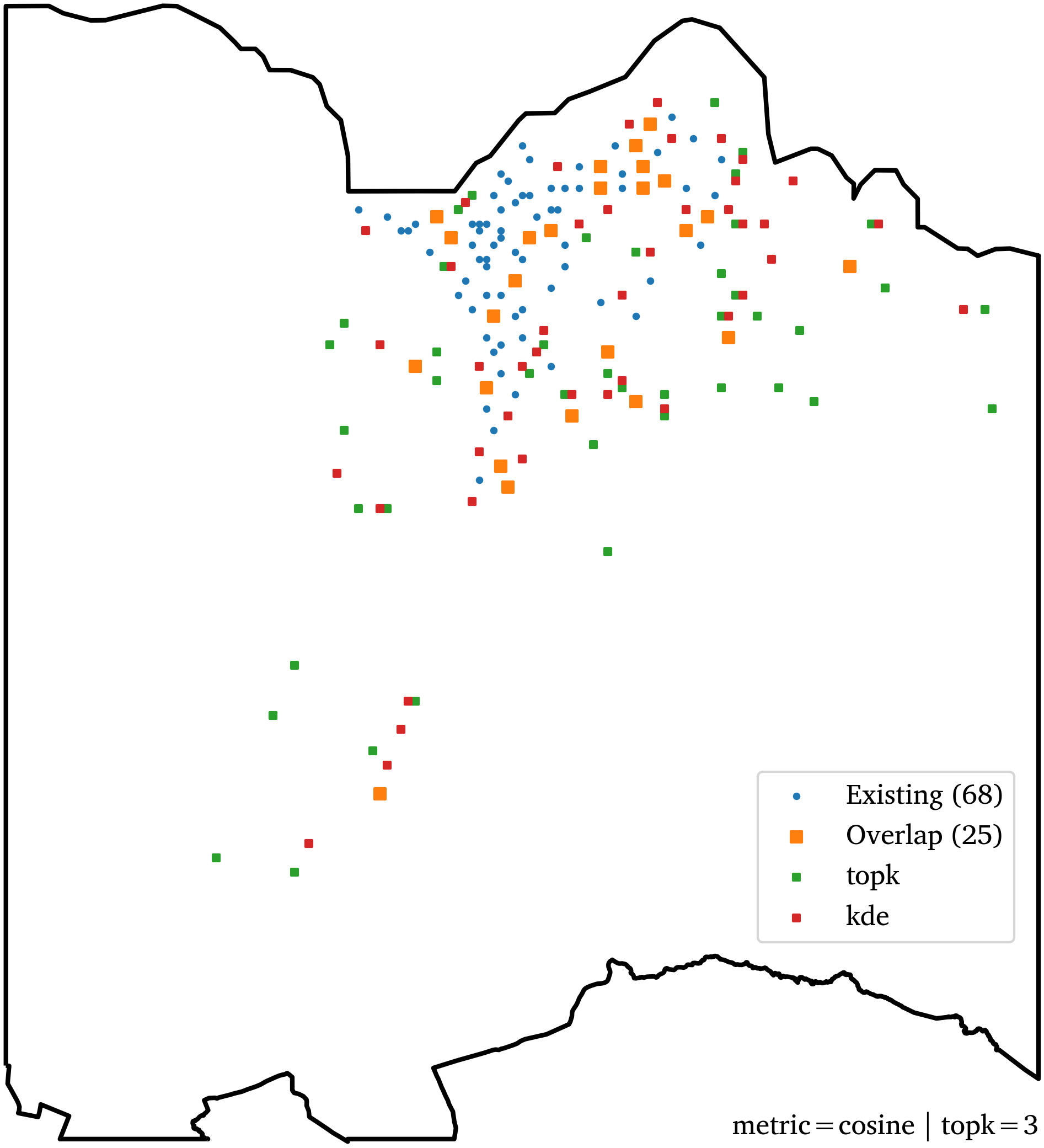}
        \caption{Methods: Top-$k$ vs.\ KDE}
        \label{fig:allocation-methods}
    \end{subfigure}
    \hfill
    \begin{subfigure}{0.49\linewidth}
        \includegraphics[width=\linewidth]{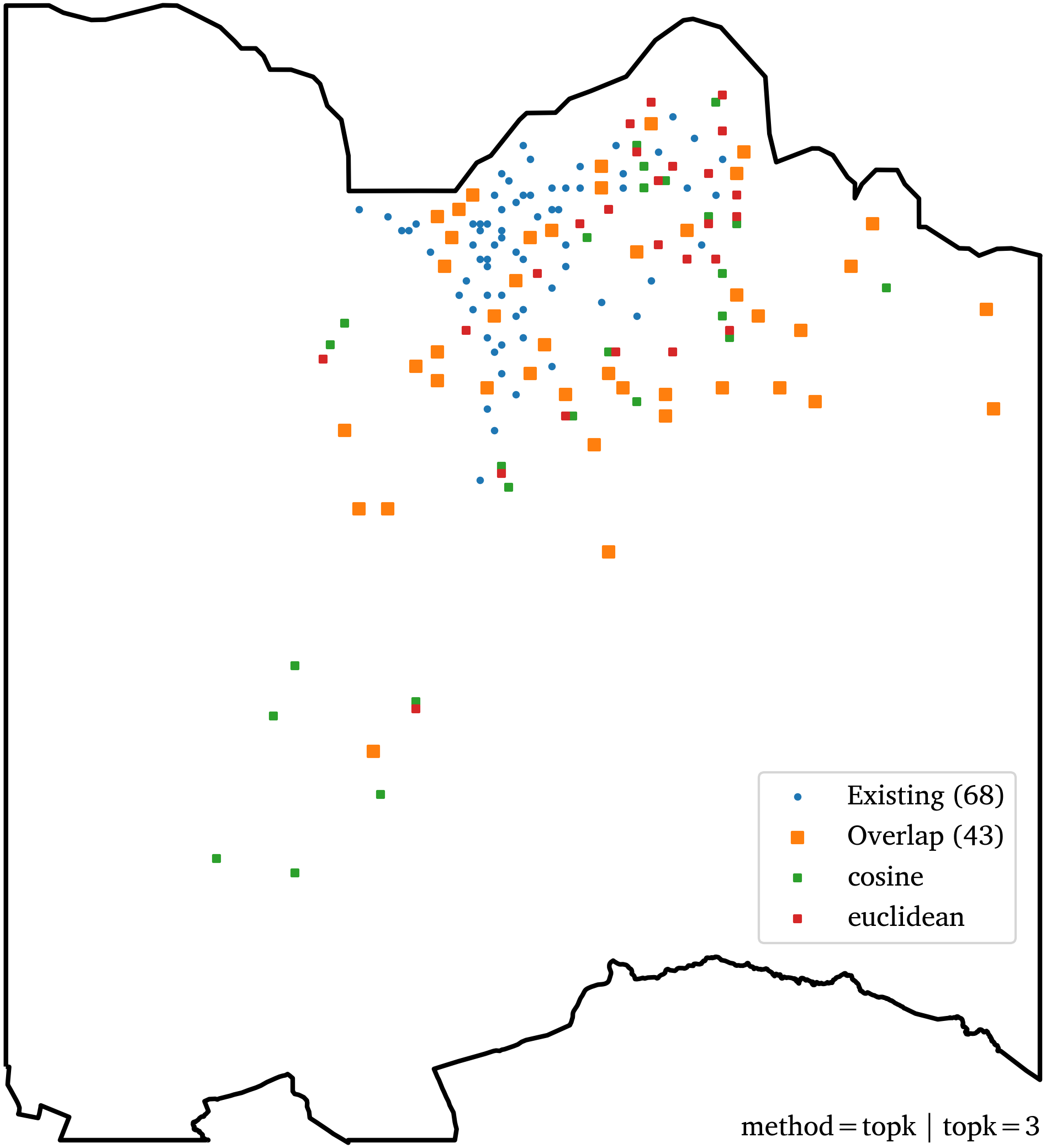}
        \caption{Metrics: Cosine vs.\ Euclidean}
        \label{fig:metric-comparison}
    \end{subfigure}
    \caption{Sensitivity of allocation outcomes to similarity aggregation method (left) and distance metric (right). Both configurations use HDAE embeddings, selecting 68 candidates with a 250m buffer.}
\end{figure*}

\vspace{0.5\baselineskip}
\subsubsection{Similarity Aggregation Method}

Fig.~\ref{fig:allocation-methods} compares spatial distributions of selected candidates under the top-$k$ and KDE similarity aggregation methods. Top-$k$ with $k=3$ produces a more spatially dispersed selection farther from existing stations, while KDE produces more conservative selections around existing station clusters and major transit corridors. There is an overlap of 25 (36.76\%) and 15 (22.06\%) candidates between both methods using the cosine and Euclidean metrics respectively.

However, as the $k$ parameter increases in the top-$k$ method, the selection converges towards the KDE outcome, with 66 (97.06\%) and 67 (97.53\%) candidates overlapping with KDE for cosine and Euclidean metrics respectively at $k=68$ (global mean). This illustrates how the top-$k$ method transitions from local similarity peaks to a more global consensus as $k$ increases, while KDE inherently captures a global similarity landscape through its kernel density estimation approach.

\vspace{0.5\baselineskip}
\subsubsection{Distance Metric}

Fig.~\ref{fig:metric-comparison} compares allocations using the cosine and Euclidean distance metrics within the top-$k$ method. The comparison is inherently asymmetric because the classification head encodes class-discriminative signal in the angular geometry of the latent space, which cosine directly exploits while Euclidean distance on z-scored embeddings mixes in magnitude information that carries no class-discriminative signal.

Despite this asymmetry, the two metrics agree substantially. At $k=3$, 43 candidates (63.23\%) overlap between the metrics, rising to 56 (82.35\%) at $k=68$ with the same overlap observed for KDE. The relationship is not strictly monotonic across intermediate $k$ values, where overlap can decrease before rising again, indicating that marginal candidates shift non-linearly as the similarity landscape transitions from local to global consensus. Overall, there is a robust core of locations identified consistently across both metrics with the choice of metric affecting only the marginal selections.

\begin{figure*}[tb!]
    \centering
    \begin{minipage}[t]{0.49\linewidth}
        \includegraphics[width=\linewidth]{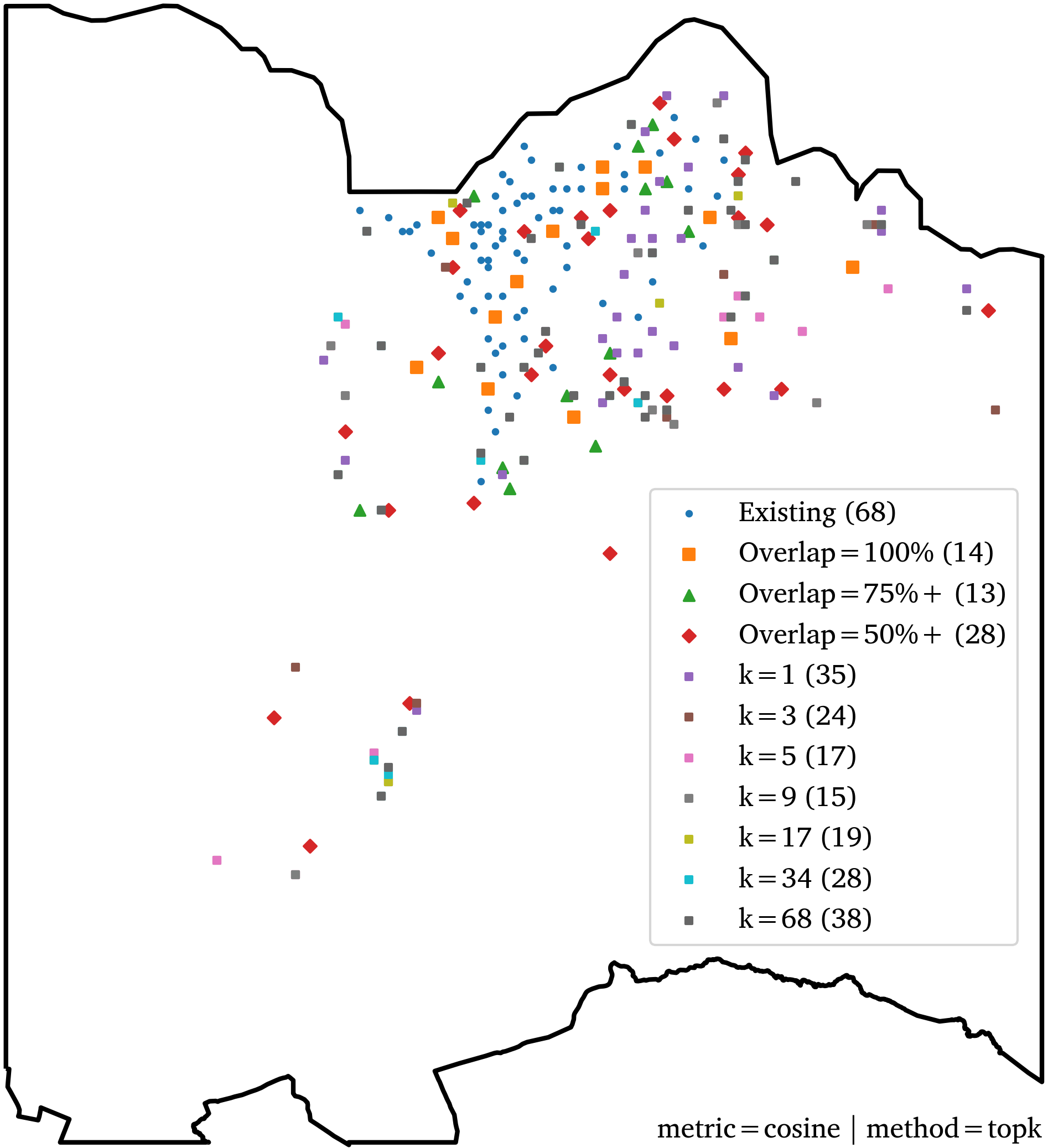}
        \caption{Sensitivity of allocation outcomes to the top-$k$ parameter across seven logarithmically-spaced values from $k=1$ (nearest) to $k=68$ (global mean).}
        \label{fig:topk-sensitivity}
    \end{minipage}
    \hfill
    \begin{minipage}[t]{0.49\linewidth}
        \includegraphics[width=\linewidth]{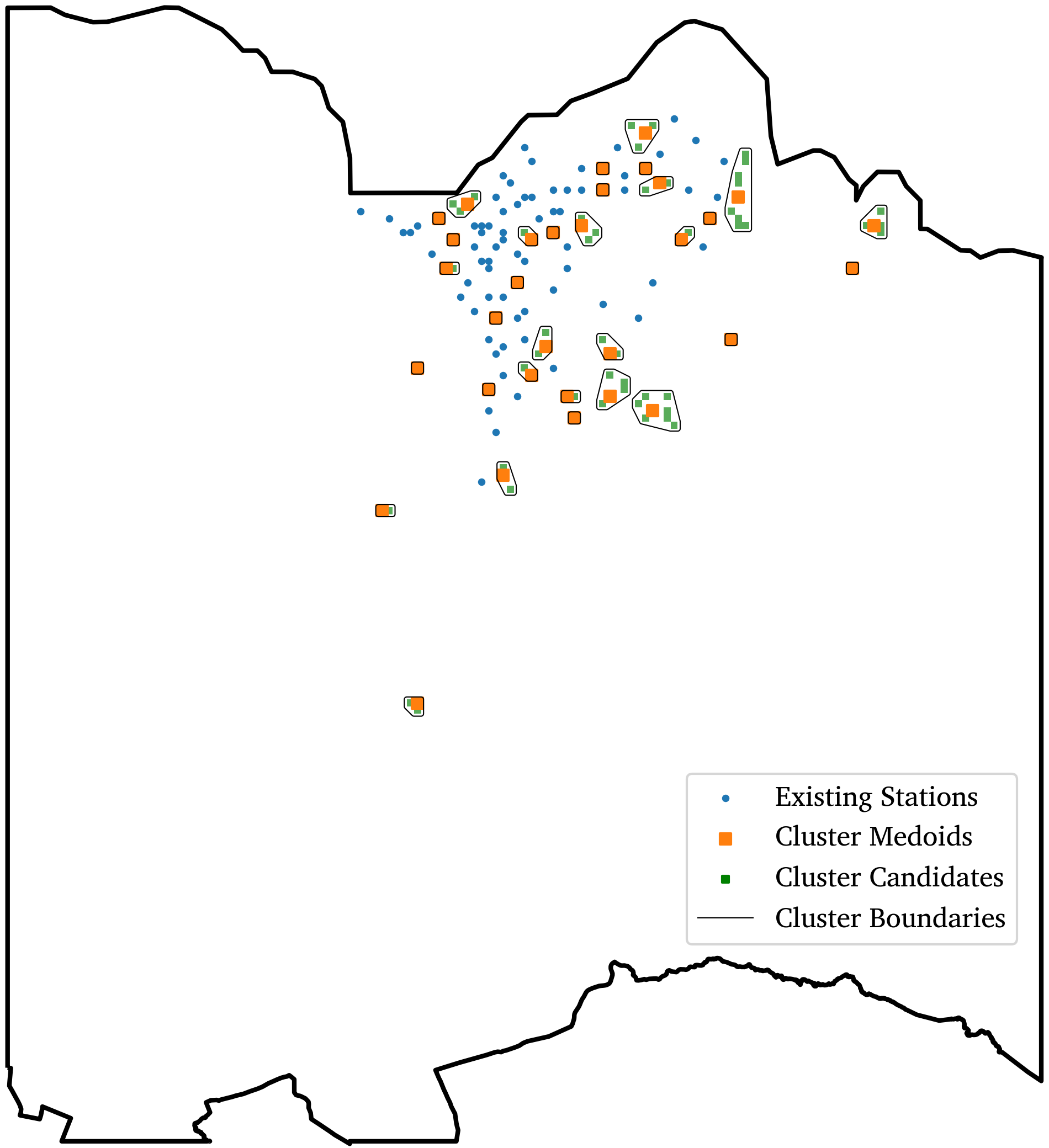}
        \caption{Consensus extension zones derived from top-$k$ sensitivity analysis, retaining clusters selected by all seven $k$ values.}
        \label{fig:consensus-extensions}
    \end{minipage}
\end{figure*}

\vspace{0.5\baselineskip}
\subsubsection{Top-$k$ Parameter}

To assess the sensitivity of the top-$k$ method to the choice of $k$, allocations are repeated across seven logarithmically-spaced values from $k=1$ (nearest neighbour) to $k=68$ (global mean) using the cosine metric: $k \in \{1, 3, 5, 9, 17, 34, 68\}$. Fig.~\ref{fig:topk-sensitivity} illustrates how the selection evolves as similarity aggregation transitions from local peaks to a global consensus.

Across all seven $k$ values, 154 unique candidates are identified out of a possible $7 \times 68 = 476$, giving a unique ratio of 2.26 compared to 7.0 if all selections were completely disjoint. This high degree of overlap reflects a stable core of locations that are consistently identified as similar to existing stations regardless of how broadly similarity is aggregated. Specifically, 14 candidates (9.09\%) are selected across all seven $k$ values, 13 (8.44\%) by at least 75\%, and 28 (18.18\%) by at least 50\%.

This stability is not uniform across the range. Adjacent large-$k$ values share more candidates than adjacent small-$k$ values, and mid-range values ($k \in \{5, 9, 17\}$) contribute disproportionately to the high-consensus set. This suggests that the most reliable similarity aggregation lies in the mid-to-global range, where selections are neither driven by isolated local peaks nor overly smoothed by the global mean.

\subsection{Consensus-Based Extension Selection}

The sensitivity analyses above demonstrate that while there is a stable core of candidates consistently identified across different similarity aggregation methods, distance metrics, and top-$k$ parameters, there are also marginal candidates that shift based on these choices. This raises the question of how to select extension locations in practice when there is no single objectively best parametrisation. To address this, we propose a consensus-based procedure that pools candidate sets generated by multiple parametrisations and retains only those locations that are consistently selected across all configurations. This removes the need to identify a single best parametrisation by instead requiring agreement across the full range of choices.

Here we demonstrate the procedure on the seven top-$k$ selections from the sensitivity analysis above. The unique candidates selected by any of the seven $k$ values are pooled together, and those within the 250m buffer from existing stations are filtered out, yielding a set of 154 unique candidate locations. Each candidate is annotated with which $k$ values selected it, allowing us to measure \it{consensus diversity}---the number of distinct $k$ values that selected each candidate. The filtered candidates are then clustered on their geographic centroids using DBSCAN \cite{Ester1996adb} with a similar neighbourhood radius of 250m and a minimum cluster size of 1 to allow for singleton clusters. This produced 89 spatial clusters, 29 with multiple candidates and 60 singletons.

The clustering step groups nearby candidates into coherent zones, and clusters are ranked by their consensus diversity with cluster size as a tie-breaker. We adopt a strict threshold by retaining only clusters achieving unanimous consensus across all seven $k$ values, yielding 32 high-confidence extension zones. Within each zone, the \it{medoid} grid (the member minimising the sum of within-cluster distances) is selected as the representative location, producing a spatially dispersed set of sites where every parametrisation agrees. Fig.~\ref{fig:consensus-extensions} visualises the consensus clusters and their selected medoids.

%%%%%=============================================================================%%%%

\section{Discussion}
\label{sec:discussion}

The HDAE architecture addresses three key challenges for similarity-based station allocation through the interplay of its components. The basic compression discards noise and redundancy in the 29-dimensional feature space, denoising augmentation ensures that grids with similar urban characteristics cluster together in the embedding space despite minor feature variations, and the classification head regularises the latent geometry so that promising candidates are embedded near the reference grids, i.e., the subset of grid cells that contain existing BSS stations. Together, these components yield substantially improved clustering quality and more spatially coherent allocation outcomes compared to raw features, which tend to produce scattered, noise-driven selections at the study area periphery.

The allocation outcomes are robust across the two independent configuration choices: the similarity aggregation method and the distance metric. The top-$k$ method at $k=3$ with cosine similarity provides a robust default, balancing sensitivity to individual reference grids with broader pattern matching. Large-$k$ top-$k$ and KDE produce nearly identical selections, confirming that both global aggregation approaches converge to similar solutions when considering the full set of reference grids. The choice of distance metric affects only marginal candidates, with a stable core of locations identified consistently across both cosine and Euclidean metrics. This robustness is partly by design, as the classification head encodes class-discriminative signal in the angular geometry of the latent space, which cosine directly exploits.

The consensus-based extension selection resolves parameter uncertainty by pooling candidates across multiple parametrisations and retaining only zones achieving unanimous agreement. We demonstrate this on the seven top-$k$ selections, clustering candidates into spatially compact zones via DBSCAN and retaining only those clusters with unanimous consensus across all seven $k$ values. This yields 32 high-confidence extension zones, each represented by a single medoid grid. This procedure applies broadly to any multi-configuration selection problem, providing a principled way to derive robust recommendations without identifying a single best parametrisation.

The framework offers significant configurability at the allocation stage. Both reference and candidate grid sets can be customised to suit planning needs without retraining the autoencoder, since similarity is computed between these defined sets. For instance, one could restrict the reference set to high-performing stations in a particular district, or limit the candidate set to a specific expansion corridor via geographic filtering. This enables targeted ``what-if'' analyses using the same learned embeddings. The framework is also purely data-driven, making no value judgement on candidate grids other than their similarity to reference grids based on engineered features. This objectivity avoids subjective biases in site selection, though it assumes the reference grids represent desirable templates worth replicating.

The similarity-based approach is therefore inherently conservative: it cannot surface locations that would succeed precisely because they differ from the current network. This can be partially mitigated by curating a reference set that represents a broader range of profiles, or by weighting reference grids by operational performance. The spatial conflict graph also relies on Euclidean buffer distances, whereas network distance or land use feasibility constraints would produce more realistic spacing rules. Finally, the core assumption that feature-space similarity predicts operational success needs to be empirically validated through longitudinal deployment studies, and the HDAE can be extended to incorporate temporal usage patterns to enable allocation strategies that adapt to seasonal demand variation.

%%%%%=============================================================================%%%%

\section{Conclusion}
\label{sec:conclusion}

This paper presented a data-driven framework for bike-sharing system (BSS) expansion that leverages a hybrid denoising autoencoder (HDAE) to learn similarity-based representations of urban characteristics without explicit demand forecasting. Demonstrated on Trondheim's bike-sharing network, the HDAE embeddings produce more spatially coherent clusters and allocation outcomes than raw features, sensitivity analyses confirm robustness across similarity methods and distance metrics, and a consensus procedure distils the results into 32 high-confidence extension zones where all parametrisations agree. The framework generalises beyond BSS to any location-allocation setting where existing desirable instances inform the selection of new candidates, and the consensus mechanism transfers to any multi-configuration selection problem requiring parameter-agnostic recommendations.

%%%%%=============================================================================%%%%

\section*{Acknowledgment}
This research received funding from the PERSEUS Doctoral Program, supported under the Marie Skłodowska-Curie grant agreement No. 101034240. The authors also acknowledge MobilitetsLab Stor-Trondheim for their financial contribution. The authors are grateful to Urban Infrastructure Partner (UIP) and AtB AS for sharing city-bike and public transit data that made this research possible.

%%%%%=============================================================================%%%%

% Generated by IEEEtran.bst, version: 1.14 (2015/08/26)

%%%%%=============================================================================%%%%

\end{document}